\documentclass[12pt]{article}

\usepackage{sbc-template}
\usepackage{graphicx,url}
\usepackage[utf8]{inputenc}
\usepackage[T1]{fontenc}  
\usepackage{amsmath}   % para ambientes matemáticos como align
\usepackage{amssymb}   % para símbolos extras, incluindo \mathbb
\usepackage{amsfonts}  % para fontes matemáticas como \mathbb
\usepackage{float}
\usepackage[nolist]{acronym}
\usepackage{tikz}
\usetikzlibrary{shapes, arrows, positioning, shadows, calc}

\usepackage{graphicx}
\usepackage{caption}
\usepackage{subcaption}
\usepackage{booktabs}
\usepackage{tikz}
\newcommand{\circlednumber}[1]{%
    \tikz[baseline=(char.base)]{
        \node[shape=circle, draw=black, fill=black, text=white, inner sep=0.2pt, font=\sffamily\small] (char) {#1};
    }
}
\usetikzlibrary{arrows.meta,positioning,shadows,calc}
\usepackage{fontawesome}
\usepackage[table]{xcolor}
\title{Evaluating Temporal and Structural Anomaly Detection Paradigms for DDoS Traffic}

\author{Yasmin Souza Lima\inst{1}, Rodrigo Moreira\inst{1}, Larissa F. {Rodrigues Moreira}\inst{1}, \\ Tereza Cristina M. de B. Carvalho\inst{3}, Flávio de Oliveira Silva\inst{2}, }

\address{Institute of Exact and Technological Sciences \\ Federal University of Viçosa
    (UFV) -- MG -- Brazil
    \nextinstitute
  Department of Informatics -- School of Engineering\\
  University of Minho (UMinho) -- Braga -- Portugal
  \nextinstitute
  University of São Paulo (USP)\\
  05.508-010 -- São Paulo -- SP -- Brazil
   \email{\{yasmin.lima, rodrigo, larissa.f.rodrigues\}@ufv.br}  
   \email{terezacarvalho@usp.br, flavio@di.uminho.pt}
}

\begin{document} 
\acrodef{3GPP}{3rd Generation Partnership Projec}
\acrodef{5G}{5th Generation Mobile Network}
\acrodef{ACF}{Autocorrelation Function}
\acrodef{AI}{Artificial Intelligence}
\acrodef{IF}{Isolation Forest}
\acrodef{LSTM}{Long Short-Term Memory}
\acrodef{OCSVM}{One-Class Support Vector Machine}
\acrodef{PCA}{Principal Component Analysis}
\acrodef{DDoS}{Distributed Denial of Service}

\maketitle

\begin{abstract}
Unsupervised anomaly detection is widely used to detect Distributed Denial-of-Service (DDoS) attacks in cloud-native 5G networks, yet most studies assume a fixed traffic representation, either temporal or structural, without validating which feature space best matches the data. We propose a lightweight decision framework that prioritizes temporal or structural features before training, using two diagnostics: lag-1 autocorrelation of an aggregated flow signal and PCA cumulative explained variance. When the probes are inconclusive, the framework reserves a hybrid option as a future fallback rather than an empirically validated branch. Experiments on two statistically distinct datasets with Isolation Forest, One-Class SVM, and KMeans show that structural features consistently match or outperform temporal ones, with the performance gap widening as temporal dependence weakens.
\end{abstract}

\section{Introduction}\label{sec:introduction}

Modern network infrastructures increasingly rely on cloud-native designs, microservice decomposition, and serverless execution models, generating large volumes of telemetry at every layer of the stack. As threat activity intensifies and \ac{AI}-enabled tooling accelerates adoption~\cite{gartner2026forecast}, the pressing question is how to turn pervasive telemetry into a representation that makes malicious behavior separable before any detector is even selected.

The main issue is not detector choice, it is how traffic is represented before training. Most prior work commits to either temporal features or structural projections such as \ac{PCA} and optimizes the detector around that choice without testing paradigm fit. Recent advances span attention based sequence models for \ac{5G} telemetry, autoencoders paired with Isolation Forest and SHAP, and clustering variants, but they still do not provide a repeatable rule to choose between temporal and structural representations in new deployments~\cite{Moreira2023}.

This paper proposes a pre-training decision framework for unsupervised \ac{DDoS} detection. It uses two probes, lag-1 autocorrelation of an aggregated flow signal and cumulative \ac{PCA} explained variance, to prioritize temporal versus structural representations before model selection. The study evaluates only these two branches on two datasets with opposing structure, one temporally dependent and one high-dimensional \ac{5G} benchmark with weak temporal signal, using \ac{IF}, \ac{OCSVM}, and KMeans. The hybrid branch is retained as a conceptual fallback for inconclusive probe combinations and is not empirically validated here. Within this scope, the results show that structural features consistently match or outperform temporal ones when autocorrelation is weak. Our contribution is therefore not a new detector, but a lightweight pre-training characterization heuristic that helps narrow the representation choice before unsupervised modelling.

\section{Related Work}\label{sec:related_work}

Unsupervised anomaly detection for network traffic and cloud-native systems has been addressed from multiple angles, yet no prior study explicitly compares temporal and structural feature paradigms or provides a decision rule for choosing between them. In the \ac{5G} and O-RAN domain, deep sequence models with attention have been used to detect abnormal Network Function interactions in the \ac{3GPP} Service Based Architecture~\cite{paper6}, Continuous Time Markov Chain models have targeted signaling anomalies via digital twins and safe reinforcement learning~\cite{paper5}, and LSTM autoencoders deployed in the Near-RT RIC have triggered secure slicing against malicious User Equipment~\cite{paper7}. For broader network traffic, \ac{IF} has been extended with X-means clustering to improve detection rates~\cite{paper11}, paired with autoencoders and SHAP-based explanations~\cite{paper8}, and combined with ConvLSTM for spatiotemporal modelling~\cite{paper9}. Complementary efforts include calibrated one-class classification under anomaly contamination~\cite{paper10}, dilated-convolution variational autoencoders for multi-scale patterns in mobile counters~\cite{paper2}, multi-modal fusion of logs and traces in microservice architectures~\cite{paper1}, and a research agenda for context-aware detection in serverless environments~\cite{paper4}. Table~\ref{tab:sota} summarises the coverage of each study across ten dimensions; columns mark whether the work targets \ac{5G}/O-RAN, fuses multiple telemetry sources, employs deep learning or specific detectors (IF, \ac{OCSVM}), includes explainability or mitigation, and, crucially, whether it compares temporal and structural paradigms or offers a paradigm decision rule.

\begin{table}[ht]
\centering
\caption{Short Related Work Comparison.}
\label{tab:sota}
\renewcommand{\arraystretch}{1.2}
\resizebox{\textwidth}{!}{%
\begin{tabular}{lcccccccccc}
\hline
\textbf{Paper}         & \textbf{\begin{tabular}[c]{@{}c@{}}5G or \\ O-RAN\end{tabular}} & \textbf{\begin{tabular}[c]{@{}c@{}}Multi-source \\ Telemetry\end{tabular}} & \textbf{\begin{tabular}[c]{@{}c@{}}Deep \\ Learning\end{tabular}} & \textbf{\begin{tabular}[c]{@{}c@{}}Uses \\ IF\end{tabular}} & \textbf{\begin{tabular}[c]{@{}c@{}}Uses \\ OCSVM\end{tabular}} & \textbf{XAI} & \textbf{Mitigation} & \textbf{\begin{tabular}[c]{@{}c@{}}Temporal vs \\ Structural\end{tabular}} & \textbf{\begin{tabular}[c]{@{}c@{}}Paradigm \\ decision rule\end{tabular}} & \textbf{\begin{tabular}[c]{@{}c@{}}Open \\ Release\end{tabular}} \\ \hline
\cite{paper1}                                      & \faCircleO                                                              & \faCircle                                                                        & \faCircle                                                               & \faCircleO                                                          & \faCircle                                                            & \faCircleO           & \faCircleO                  & \faCircleO                                                                         & \faCircleO                                                                         & \faCircleO                                                               \\
\cite{paper2}                                      & \faCircleO                                                              & \faCircleO                                                                         & \faCircle                                                               & \faCircleO                                                          & \faCircleO                                                             & \faCircleO           & \faCircleO                  & \faCircleO                                                                         & \faCircleO                                                                         & \faCircle                                                              \\
\cite{paper4}                                     & \faCircleO                                                              & \faCircleO                                                                         & \faCircleO                                                                & \faCircleO                                                          & \faCircleO                                                             & \faCircleO           & \faCircleO                  & \faCircleO                                                                         & \faCircleO                                                                         & \faCircleO                                                               \\
\cite{paper5}                                     & \faCircle                                                             & \faCircle                                                                        & \faCircle                                                               & \faCircleO                                                          & \faCircleO                                                             & \faCircleO           & \faCircle                 & \faCircleO                                                                         & \faCircleO                                                                         & \faCircleO                                                               \\
\cite{paper6}                                     & \faCircle                                                             & \faCircleO                                                                         & \faCircle                                                               & \faCircleO                                                          & \faCircleO                                                             & \faCircleO           & \faCircleO                  & \faCircleO                                                                         & \faCircleO                                                                         & \faCircleO                                                               \\
\cite{paper7}                                      & \faCircle                                                             & \faCircleO                                                                         & \faCircle                                                               & \faCircleO                                                          & \faCircleO                                                             & \faCircleO           & \faCircle                 & \faCircleO                                                                         & \faCircleO                                                                         & \faCircleO                                                               \\
\cite{paper8}                                     & \faCircleO                                                              & \faCircleO                                                                         & \faCircle                                                               & \faCircle                                                         & \faCircleO                                                             & \faCircle          & \faCircleO                  & \faCircleO                                                                         & \faCircleO                                                                         & \faCircleO                                                               \\
\cite{paper9}                                      & \faCircleO                                                              & \faCircleO                                                                         & \faCircle                                                               & \faCircle                                                         & \faCircleO                                                             & \faCircle          & \faCircleO                  & \faCircleO                                                                         & \faCircleO                                                                         & \faCircleO                                                               \\
\cite{paper10}                                   & \faCircleO                                                              & \faCircleO                                                                         & \faCircle                                                               & \faCircleO                                                          & \faCircleO                                                             & \faCircleO           & \faCircleO                  & \faCircleO                                                                         & \faCircleO                                                                         & \faCircle                                                              \\
\cite{paper11}                                    & \faCircleO                                                              & \faCircleO                                                                         & \faCircleO                                                                & \faCircle                                                         & \faCircleO                                                             & \faCircleO           & \faCircleO                  & \faCircleO                                                                         & \faCircleO                                                                         & \faCircleO                                                               \\ \hline
\textbf{Our Approach}                  & \faCircle                                                             & \faCircleO                                                                         & \faCircleO                                                                & \faCircle                                                         & \faCircle                                                            & \faCircleO           & \faCircleO                  & \faCircle                                                                        & \faCircle                                                                        & \faCircle                                                               \\ \hline
\end{tabular}%
}
\end{table}

\textbf{Contribution Positioning.} Unlike prior detector-specific studies, this paper addresses the paradigm choice by comparing temporal and structural representations for unsupervised \ac{DDoS} detection across statistically diverse datasets. Our conceptual contribution is a lightweight decision procedure for unseen deployments using two diagnostic probes: lag-1 autocorrelation and \ac{PCA} cumulative explained variance. This framework bridges detector-centric research and deployment-time decision-making, particularly in \ac{5G} environments where optimal traffic representation fluctuates.

\section{Methodology}\label{sec:methodology}

\begin{figure}[ht]
\centering
\resizebox{\linewidth}{!}{%
\begin{tikzpicture}[
    font=\sffamily,
    >=Latex,
    node distance=12mm and 18mm,
    arrow/.style={-Latex, very thick, draw=black!75},
    yes/.style={arrow, draw=green!55!black},
    no/.style={arrow, draw=red!65!black},
    tag/.style={font=\bfseries\footnotesize, inner sep=1.5pt, fill=white, fill opacity=0.92, text opacity=1},
    box/.style={draw, rounded corners=3pt, align=center, text width=4.2cm,
                minimum height=10mm, inner sep=5pt, drop shadow, fill=#1},
    decision/.style={draw, rounded corners=3pt, align=center, text width=4.2cm,
                     minimum height=10mm, inner sep=5pt, drop shadow, fill=blue!12},
    outcome/.style={draw, rounded corners=4pt, align=center, text width=5.4cm,
                    minimum height=12mm, inner sep=6pt, drop shadow, fill=#1, font=\bfseries}
]

% Main left-to-right spine
\node[box=gray!12] (input) {Input, network traffic dataset};
\node[box=yellow!15, right=of input] (acf) {Compute ACF at lag 1\\on an aggregated flow signal};
\node[decision, right=of acf] (d1) {Is lag-1 ACF high?};
\node[box=yellow!15, right=of d1, xshift=10mm] (pca) {Compute PCA cumulative\\explained variance};
\node[decision, right=of pca] (d2) {$\geq 95\%$ variance with $\leq 5$ PCs, ?};

% Outcomes (kept close to reduce height)
\node[outcome=cyan!18, above=16mm of d1] (temporal) {Temporal paradigm\\One Class SVM on rolling features\\Isolation Forest in temporal space};

\node[outcome=green!18, above=16mm of d2] (hybrid) {Hybrid paradigm\\Combine temporal and structural features};

\node[outcome=red!12, below=16mm of d2] (structural) {Structural paradigm\\K Means on PCA space\\Isolation Forest on structural features};

% Links
\draw[arrow] (input) -- (acf);
\draw[arrow] (acf) -- (d1);
\draw[no]    (d1) -- node[tag, above]{No} (pca);
\draw[arrow] (pca) -- (d2);

\draw[yes] (d1.north) -- ++(0,7mm) -| node[tag, near start, above]{Yes} (temporal.west);

\draw[no]  (d2.north) -- ++(0,7mm) -| node[tag, near start, above]{No} (hybrid.west);

\draw[yes] (d2.south) -- ++(0,-7mm) -| node[tag, near start, below]{Yes} (structural.west);

\end{tikzpicture}%
}
\caption{Lightweight framework for prioritizing temporal or structural representations.}
\label{fig:pipeline}
\end{figure}
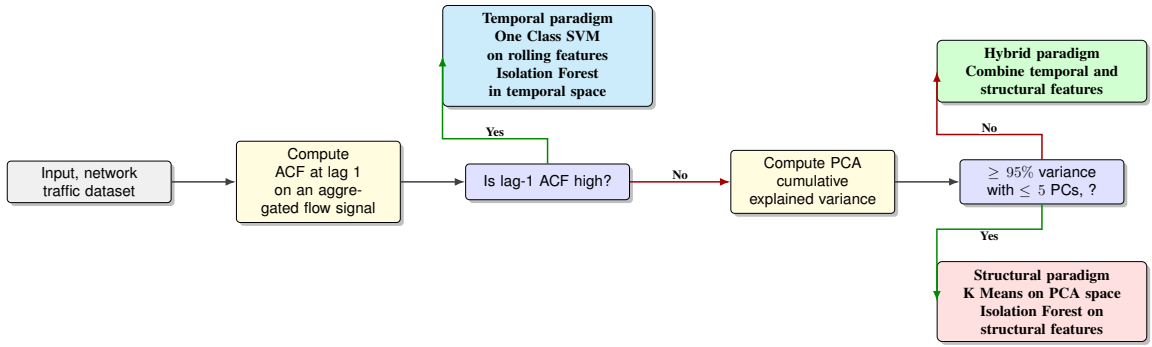

The pipeline consists of six sequential steps, illustrated in Figure~\ref{fig:pipeline}: \circlednumber{1}~dataset selection and characterization, \circlednumber{2}~preprocessing and cleaning, \circlednumber{3}~feature engineering under two paradigms, \circlednumber{4}~unsupervised detection, \circlednumber{5}~evaluation against ground-truth labels, and \circlednumber{6}~cross-paradigm
comparison.

% ---- Figure stays unchanged ----
% Requires:
% \usepackage{tikz}
% \usetikzlibrary{arrows.meta,positioning,shadows,calc}

\subsection{Datasets}\label{sec:datasets}

\textbf{Step~\circlednumber{1}.}
Two labelled datasets with distinct traffic profiles were selected to
evaluate whether detection paradigm effectiveness depends on the
statistical nature of the traffic.

\textbf{CICDDoS2019.}
The CICDDoS2019 benchmark~\cite{sharafaldin2019} contains bidirectional
flow records captured during simulated \ac{DDoS} campaigns.  Each record
includes 80 features extracted by CICFlowMeter, covering packet counts,
byte volumes, inter-arrival times, and flag distributions.  The dataset
includes eight reflection-based attack types, each stored as a separate
CSV file.  A stratified sample of 30\,000 flows per file was drawn,
yielding approximately 240\,000 records with a heavily imbalanced class
distribution (benign flows constitute less than 0.4\%).

\textbf{5GAD.}
The 5G Attack Detection dataset (5GAD)~\cite{Coldwell2022} provides simulated
traffic from a 5G network slicing testbed.  Records are fixed-length
feature vectors of 1\,024 dimensions stored as NumPy arrays.  Labels
are binary (\texttt{normal}/\texttt{attack}), evenly distributed
across 48\,320 samples.  The original tensor shape was flattened to a
standard two-dimensional matrix prior to analysis.

\subsection{Preprocessing}\label{sec:preproc}

\textbf{Step~\circlednumber{2}.}
All experiments ran on an Intel Core~i7 workstation with 32\,GB RAM
using Python~3.12 and scikit-learn~1.4.  Numeric precision was reduced
to \texttt{float32} and the \ac{OCSVM} prediction phase was batched in
chunks of 20\,000 samples to fit memory.

Both datasets underwent a uniform cleaning procedure: non-numeric
columns were removed from CICDDoS2019, infinite values and
\texttt{NaN} entries were replaced with zero, and all features were
standardised to zero mean and unit variance via z-score normalisation.
This ensures that distance-based algorithms are not biased by scale
differences across features.

\subsection{Feature Engineering}\label{sec:features}

\textbf{Step~\circlednumber{3}.}
Two independent feature spaces were constructed from the same standardised input.

\textbf{Temporal feature space.}
Rolling statistics were computed over windows of size
$w \in \{10, 30, 100\}$ along the sample-index axis, preserving the
sequential ordering of the original capture.  Six statistics were
extracted per window per base signal (rolling mean, standard deviation,
maximum, minimum, first-order difference, and coefficient of variation),
applied to the $\ell_2$ norm and the first five principal components,
yielding $3 \times 6 \times 6 = 108$ temporal features.  The
coefficient of variation is defined as Eq.~\ref{eq:cv}:
\begin{equation}\label{eq:cv}
    \text{CV}_{w,i} =
    \begin{cases}
        \dfrac{\hat{\sigma}_{w,i}}{|\hat{\mu}_{w,i}|} & \text{if }
        |\hat{\mu}_{w,i}| > \epsilon, \\[6pt]
        0 & \text{otherwise},
    \end{cases}
\end{equation}
where $\hat{\mu}_{w,i}$ and $\hat{\sigma}_{w,i}$ are the rolling mean and standard deviation within the window centred at sample~$i$, and $\epsilon = 10^{-8}$.

\textbf{Structural feature space.}
\ac{PCA} was applied to the standardised data, retaining the first $d{=}10$ components.  Each sample is projected independently onto the leading eigenvectors of the covariance matrix (Eq.~\ref{eq:pca}):
\begin{equation}\label{eq:pca}
    \mathbf{z}_i = \mathbf{W}^\top \tilde{\mathbf{x}}_i,
    \quad \mathbf{W} \in \mathbb{R}^{p \times d},
\end{equation}
embedding no temporal context.

\subsection{Unsupervised Detection Methods}\label{sec:methods}

\textbf{Step~\circlednumber{4}.}
We evaluated eight experimental configurations across two datasets. Isolation Forest was applied in both temporal and structural spaces, One Class SVM was evaluated in the temporal space, and KMeans was evaluated in the structural space.

\textbf{Isolation Forest (IF).}
IF isolates anomalies via recursive random partitioning; samples requiring fewer splits receive higher anomaly scores. For numerical stability under extreme class imbalance, the contamination parameter was set as
$c = \min(\max(\hat{r},\, 0.01),\, 0.35)$, where $\hat{r}$ is the dataset-level attack ratio. This uses label information available only in the offline benchmark and should therefore be interpreted as evaluation-time calibration rather than a deployment-ready setting.

\textbf{One-Class SVM (OCSVM).}
\ac{OCSVM} learns a boundary around normal traffic in kernel space.  An RBF kernel with $\gamma{=}\text{scale}$ was used, trained on the first 5\,000 samples of each dataset under a cold-start assumption, with $\nu = \min(c,\, 0.3)$. This protocol assumes that the initial traffic segment is a reasonable proxy for nominal behaviour, which is a pragmatic benchmark choice rather than a universally valid operational assumption.

\textbf{KMeans.}
KMeans partitions the feature space into
$K{=}2$ clusters, consistent with the dominant binary regime and the peak Silhouette behaviour observed in Figure~\ref{fig:silhouette}. After convergence, each cluster is associated post hoc with the majority ground-truth label among its members for external evaluation only; the clustering step itself remains fully unsupervised.

\subsection{Evaluation and Cross-Paradigm Comparison}\label{sec:evaluation}

\textbf{Steps~\circlednumber{5}--\circlednumber{6}.}
Detection performance was assessed using Precision~($P$), Recall~($R$),
and F1-Score against ground-truth labels, with the attack class as the
positive class.  For KMeans, clustering quality was additionally measured
by the Silhouette Score.

The paradigm gap for each metric $m \in \{P, R, F1\}$ quantifies the difference between the best temporal and the best structural method (Eq.~\ref{eq:gap}).
\begin{equation}\label{eq:gap}
    \Delta_m = \max_{t \in \mathcal{T}} m_t
             - \max_{s \in \mathcal{S}} m_s,
\end{equation}
where $\mathcal{T}$ and $\mathcal{S}$ are the temporal and structural method sets. Negative values indicate structural superiority. For unseen deployments, we interpret two diagnostic probes, namely lag-1 autocorrelation of the aggregated flow signal and cumulative variance explained by the first five principal components, as a lightweight characterization heuristic rather than a fully validated decision rule. In this paper, these probes are used primarily as a lightweight characterization heuristic to contrast strong versus weak temporal dependence and compact versus diffuse structural geometry.

\section{Results and Discussion}\label{sec:results}

We analyse how temporal dependence and feature space geometry shape the effectiveness of unsupervised \ac{DDoS} detection.  We compare temporal and structural pipelines on CICDDoS2019 and 5GAD, using the figures to extract four empirical insights.

\begin{figure}[ht]
  \centering
  % --- FIGURA DA ESQUERDA (ACF Comparison) ---
  \begin{minipage}{0.48\textwidth}
    \centering
    \includegraphics[width=\linewidth]{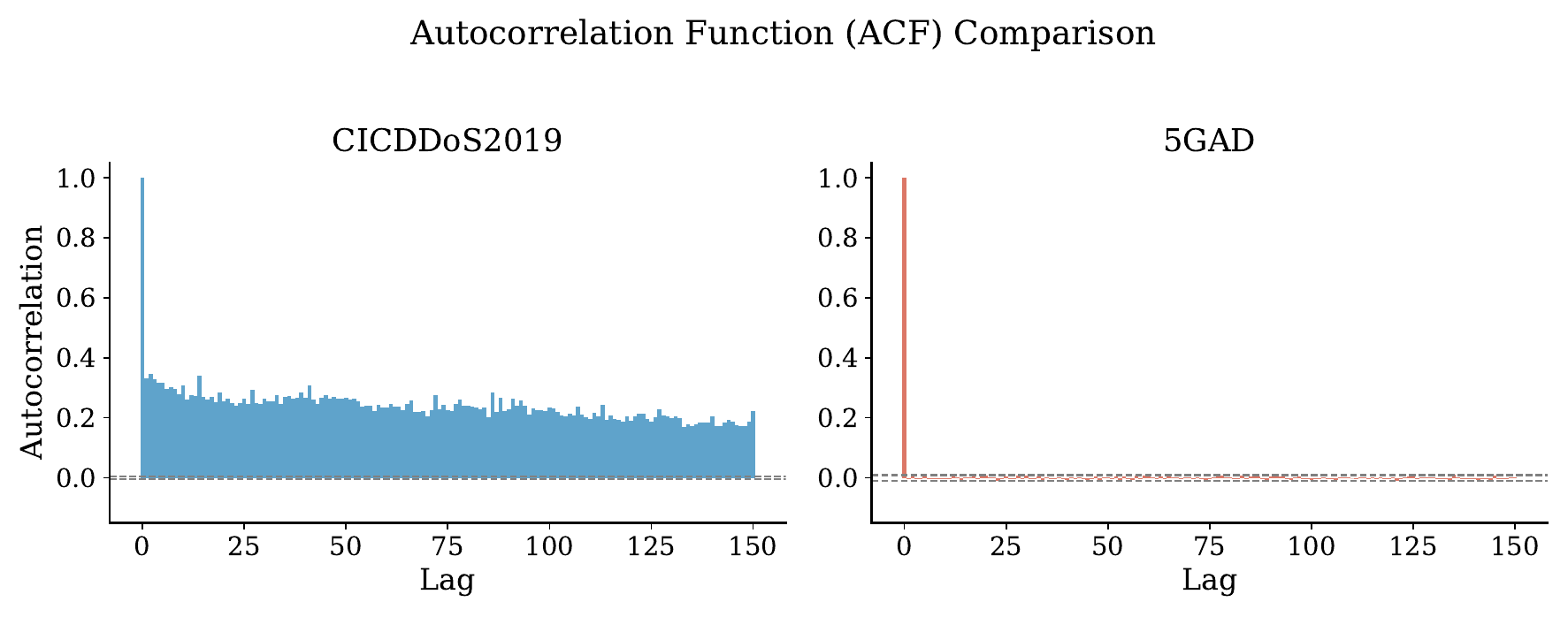}
    \caption{ACF of the aggregated flow signal for CICDDoS2019 (left) and 5GAD (right).}
    \label{fig:acf}
  \end{minipage}
  \hfill % Espaço elástico entre as figuras
  % --- FIGURA DA DIREITA (PCA Variance) ---
  \begin{minipage}{0.48\textwidth}
    \centering
    \includegraphics[width=\linewidth]{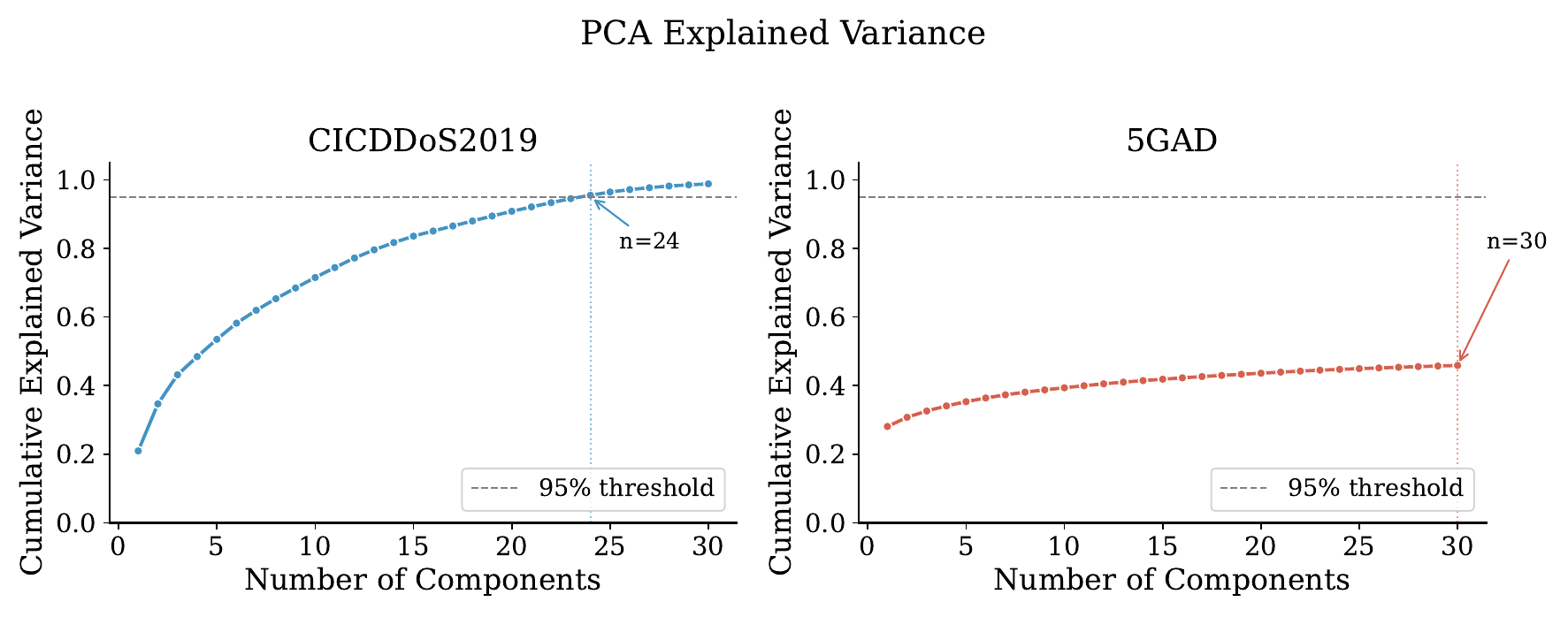}
    \caption{Cumulative \ac{PCA} explained variance for both datasets.}
    \label{fig:pca_var}
  \end{minipage}
\end{figure}

%% ── Insight 1 ──────────────────────────────────────────────
\textbf{Temporal dependence separates when temporal features can help.}
Figure~\ref{fig:acf} shows that CICDDoS2019 preserves sequential structure across lags, while 5GAD behaves as near independent samples beyond lag zero.  Figure~\ref{fig:pca_var} reinforces this contrast, CICDDoS2019 is more compressible under \ac{PCA}, while 5GAD spreads variance across many directions.  Together, these probes indicate that temporal rollups can encode meaningful context in CICDDoS2019, whereas 5GAD offers little sequential signal to exploit.

\begin{figure}[htbp]
  \centering
  % --- FIGURA DA ESQUERDA (PCA) ---
  \begin{minipage}{0.48\textwidth}
    \centering
    \includegraphics[width=\linewidth]{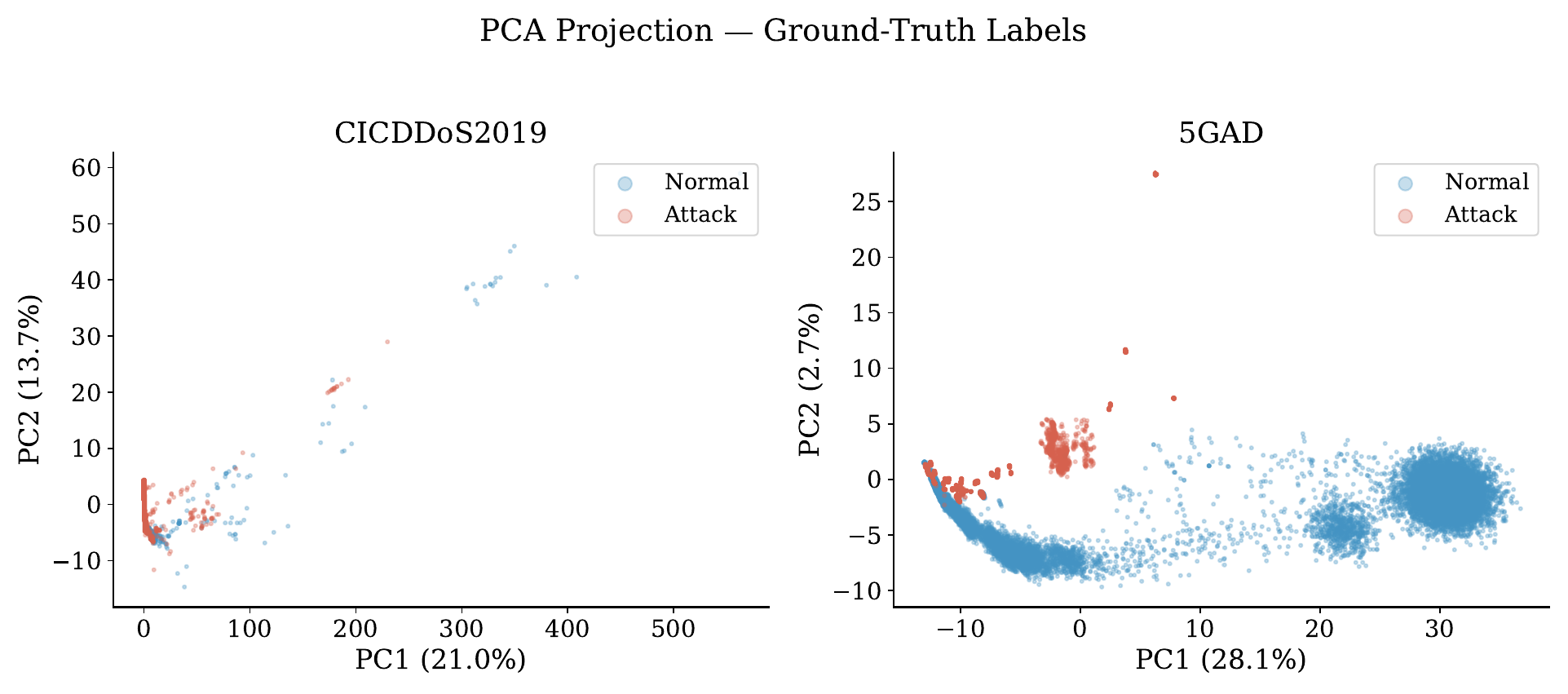}
    \caption{PCA projection coloured by ground truth labels.}
    \label{fig:pca_scatter}
  \end{minipage}
  \hfill % Empurra a outra figura para a extremidade direita
  % --- FIGURA DA DIREITA (Silhouette) ---
  \begin{minipage}{0.48\textwidth}
    \centering
    \includegraphics[width=\linewidth]{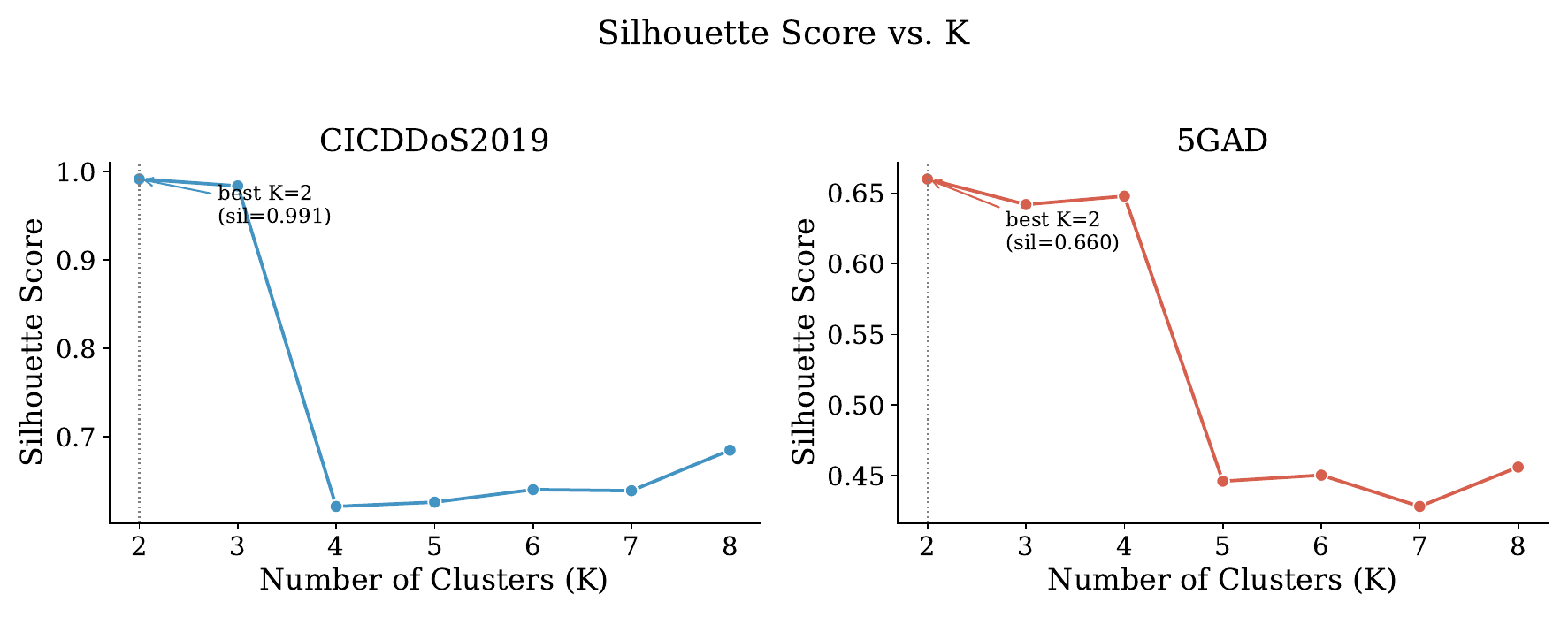}
    \caption{Silhouette Score as a function of $K$ for both datasets.}
    \label{fig:silhouette}
  \end{minipage}
\end{figure}

%% ── Insight 2 ──────────────────────────────────────────────
\textbf{Both datasets contain spatial structure, but separability differs.}
In Figure~\ref{fig:pca_scatter}, CICDDoS2019 shows stronger overlap between
classes in the leading projection, suggesting that proximity alone is not
a reliable discriminator, while 5GAD exhibits clearer separation patterns
that clustering can exploit.  Figure~\ref{fig:silhouette} is consistent
with this picture, low $K$ captures the dominant regimes, and 5GAD retains
moderate structure for additional partitions, indicating sub profiles in
attack traffic.

\begin{figure}[htbp]
  \centering
  % --- FIGURA DA ESQUERDA (Barplot) ---
  \begin{minipage}{0.48\textwidth}
    \centering
    \includegraphics[width=\linewidth]{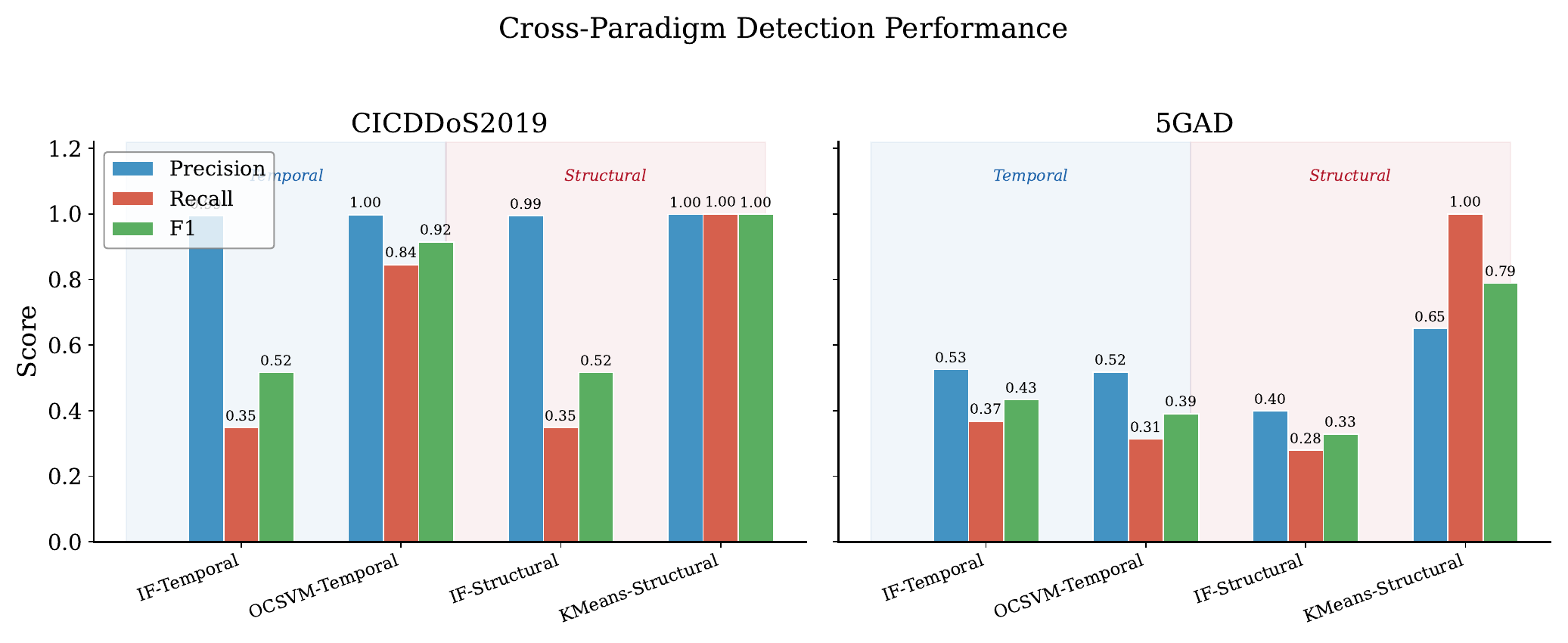}
    \caption{Detection performance for temporal and structural methods.}
    \label{fig:barplot}
  \end{minipage}
  \hfill
  % --- FIGURA DA DIREITA (Radar) ---
  \begin{minipage}{0.48\textwidth}
    \centering
    \includegraphics[width=\linewidth]{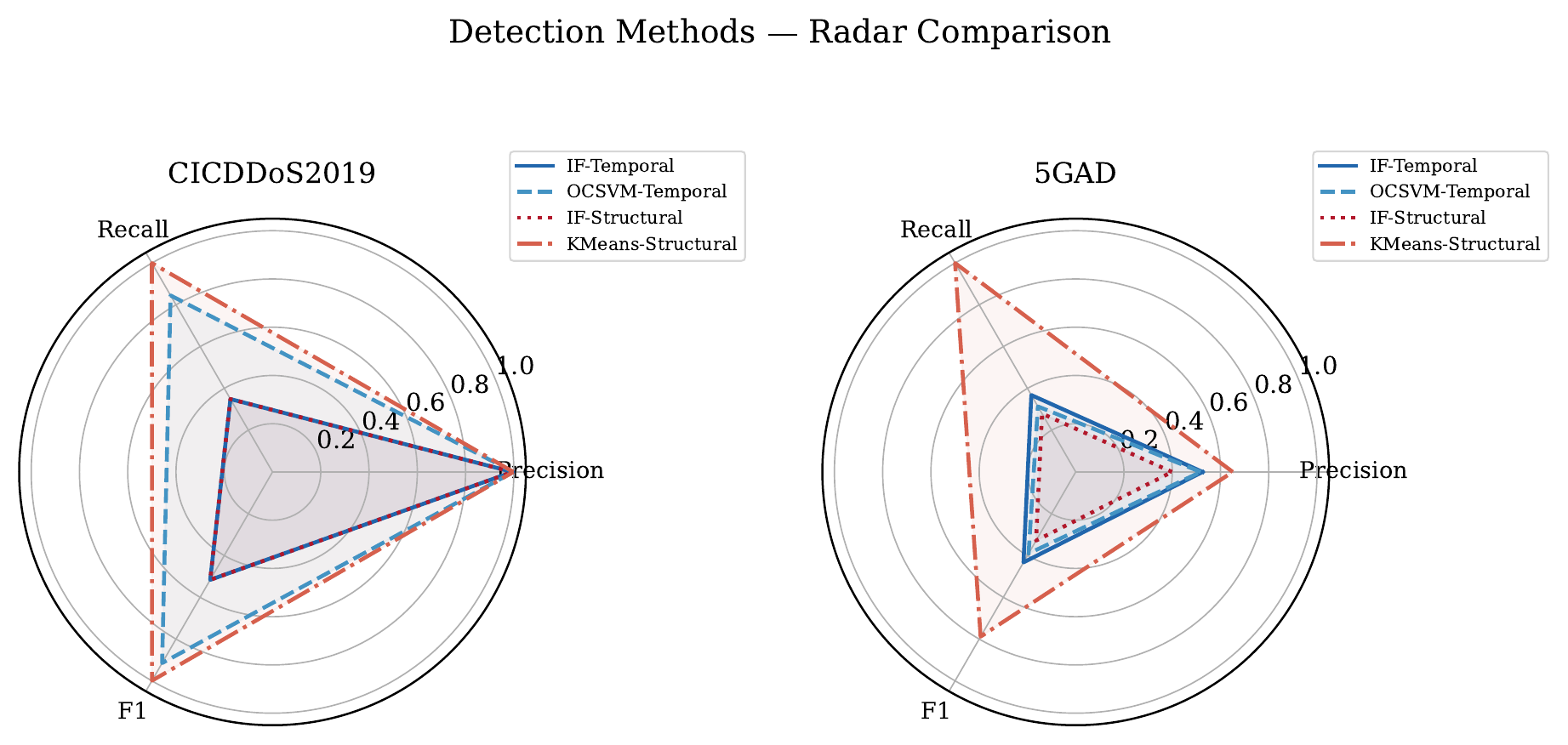}
    \caption{Radar comparison of detection methods on both datasets.}
    \label{fig:radar}
  \end{minipage}
\end{figure}

%% ── Insight 3 ──────────────────────────────────────────────
\textbf{Structural KMeans is the strongest overall, but temporal methods can be competitive when autocorrelation exists.}
Figure~\ref{fig:barplot} shows that KMeans in the structural space
achieves the best balance across metrics on both datasets.  On CICDDoS2019,
the best temporal method remains close, which aligns with the strong
sequential dependence in Figure~\ref{fig:acf}.  On 5GAD, temporal methods
lose effectiveness, matching the lack of exploitable temporal structure.
Figure~\ref{fig:radar} summarises this shift, temporal polygons shrink on
5GAD, while the structural method retains coverage.

\begin{figure}[htbp]
  \centering
  % --- LADO ESQUERDO: FIGURA ---
  \begin{minipage}{0.49\textwidth}
    \centering
    \includegraphics[width=\linewidth]{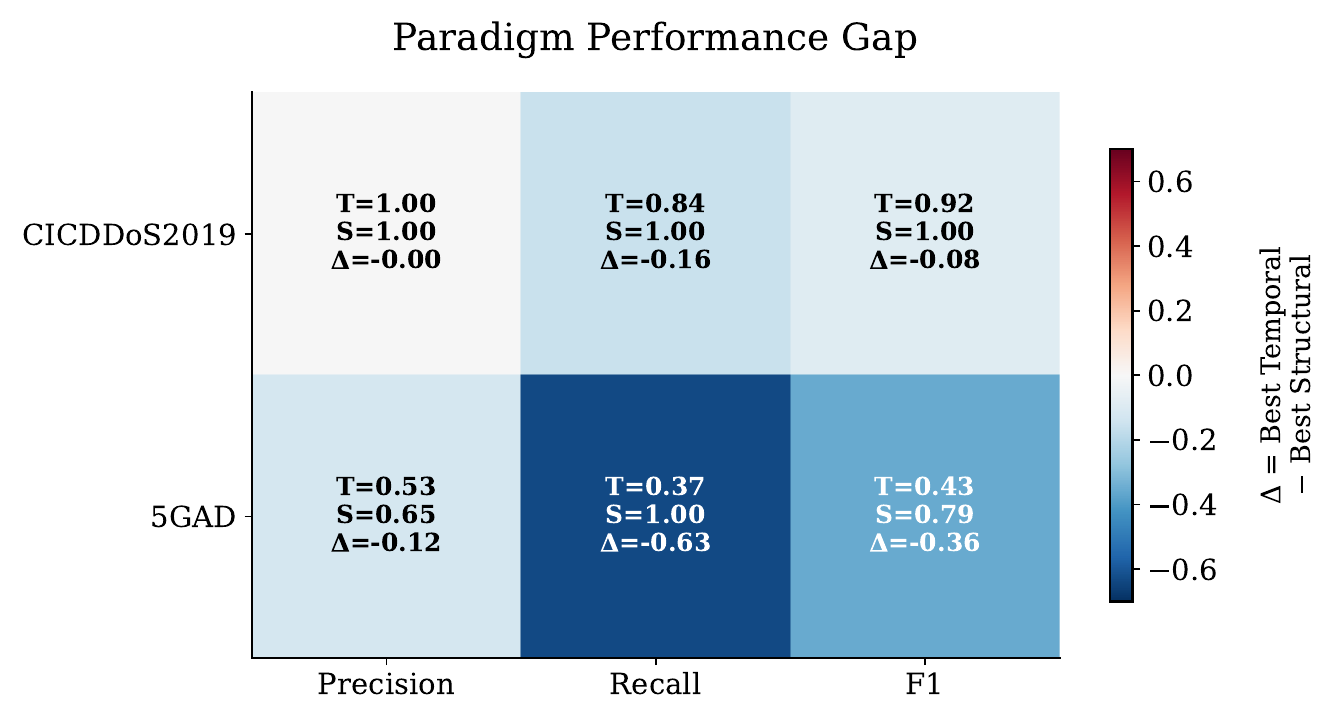}
    \caption{Paradigm performance gap, best temporal minus best structural.}
    \label{fig:gap}
  \end{minipage}
  \hfill % Espaço elástico entre os dois
  % --- LADO DIREITO: TABELA ---
  \begin{minipage}{0.50\textwidth}
    \centering
    \scriptsize
    \begin{tabular}{lcccr}
      \toprule
      \textbf{Method} & \textbf{Prec.} & \textbf{Rec.} & \textbf{F1} & \textbf{Time} \\
      \midrule
      \multicolumn{5}{l}{\textbf{Dataset: CICDDoS2019}} \\
      KMeans-Str.  & 0.998 & 1.000 & 0.999 & 1.24 \\
      OCSVM-Temp.  & 0.998 & 0.845 & 0.915 & 26.30 \\
      IF-Temp.     & 0.995 & 0.349 & 0.517 & 2.34 \\
      IF-Str.      & 0.995 & 0.349 & 0.517 & 1.65 \\
      \midrule
      \multicolumn{5}{l}{\textbf{Dataset: 5GAD}} \\
      KMeans-Str.  & 0.651 & 1.000 & 0.788 & 0.96 \\
      IF-Temp.     & 0.526 & 0.368 & 0.433 & 0.77 \\
      OCSVM-Temp.  & 0.518 & 0.314 & 0.391 & 5.95 \\
      IF-Str.      & 0.399 & 0.279 & 0.329 & 0.72 \\
      \bottomrule
    \end{tabular}
    \captionof{table}{Cross-paradigm detection results.}
    \label{tab:results}
  \end{minipage}
\end{figure}

%% ── Insight 4 ──────────────────────────────────────────────
\textbf{Structural methods never underperform the best temporal alternative, and their advantage grows when temporal structure vanishes.}
Figure~\ref{fig:gap} aggregates the paradigm gap and shows that the best
structural configuration dominates across metrics on both datasets.  The
margin is smallest when autocorrelation is strong and temporal rollups are
informative, and it increases when samples are effectively independent.
This supports the characterization logic embedded in the framework. An \ac{ACF} probe helps indicate whether temporal context is likely to be informative, while \ac{PCA} compressibility helps assess whether a compact structural projection may suffice.

\textbf{Operationally, structural pipelines provide the best cost-benefit tradeoff.}
Table~\ref{tab:results} shows that structural KMeans is consistently efficient and highly effective. Temporal \ac{OCSVM} can also perform well when traffic has strong temporal redundancy, but its much higher runtime limits real-time use. Isolation Forest is lightweight in both spaces, yet it is less reliable in the most imbalanced setting, so it is better used as a fast baseline than as the main option when structural clustering is available. Two cautions apply. Near-perfect results on CICDDoS2019 likely reflect strong benchmark separability and should not be directly generalized to operational traffic. Also, the IF contamination setting, the cold-start training slice for \ac{OCSVM}, and the post hoc majority-label mapping for KMeans were used to stabilize offline benchmarking, not to define a fully label-free deployment recipe.

%\textbf{Operationally, structural pipelines offer the best cost to benefit trade off.} Table~\ref{tab:results} indicates that KMeans in the structural space is consistently efficient while remaining highly effective.  Temporal \ac{OCSVM} can be competitive when the traffic is temporally redundant, but it carries a substantially higher runtime, which is a practical constraint for real time monitoring.  Isolation Forest is lightweight in both spaces, but it is less reliable in the most imbalanced regime, so in deployment it is best treated as a fast baseline rather than the primary choice when clustering structure is available. Two cautions are needed when interpreting these results. First, the near-perfect scores on CICDDoS2019 likely reflect strong benchmark separability and should not be extrapolated directly to operational traffic. Second, the IF contamination setting, the cold-start training slice for \ac{OCSVM}, and the post hoc majority-label mapping used for KMeans were adopted to stabilize offline benchmarking, not to claim a fully label-free deployment recipe.

\section{Concluding Remarks}\label{sec:conclusion}

We treat unsupervised \ac{DDoS} detection as a paradigm selection problem. Structural features, KMeans on \ac{PCA}, match or beat temporal alternatives, especially when autocorrelation is weak. Temporal \ac{OCSVM} helps only when autocorrelation is strong. We therefore suggest a two-probe characterization heuristic based on lag-1 autocorrelation and \ac{PCA} explained variance to guide representation choice before training. The evaluation relied on labelled benchmarks with known class structure, and the hybrid branch of the decision framework was not empirically validated. Extending the analysis to unlabelled production traces, calibrating the autocorrelation threshold across broader network environments, and integrating the procedure into an online pipeline that re-evaluates paradigm fit under distribution shift are natural next steps.

\section*{Acknowledgment}

The authors thank FAPEMIG (Grant \#APQ00923-24), FAPESP MCTIC/CGI Research project 2018/23097-3 - SFI2 - Slicing Future Internet Infrastructures, Coordenação de Aperfeiçoamento de Pessoal de Nível Superior - Brasil (CAPES) - Finance Code 001 for supporting this work.

\bibliographystyle{sbc}
\bibliography{references}

\end{document}